\documentclass[11pt]{article}

\usepackage[preprint]{acl}

\usepackage{times}
\usepackage{latexsym}
\usepackage[T1]{fontenc}
\usepackage[utf8]{inputenc}
\usepackage{microtype}
\usepackage{inconsolata}
\usepackage{graphicx}
\usepackage{afterpage}
\usepackage{colortbl}
\usepackage{soul}
\usepackage{array}
\usepackage{multirow}
\usepackage{amsmath}
\usepackage{amsfonts}
\usepackage{ulem}
\usepackage{algorithm} 
\usepackage{multicol}
\usepackage{enumitem}
\usepackage{amsthm}

\usepackage{setspace}
\usepackage{booktabs}
\usepackage{algpseudocode}
\usepackage{subcaption} 
\usepackage{mdframed} 
\usepackage{tabularx}

\title{Dual Consensus: Escaping from Spurious Majority in Unsupervised RLVR via Two-Stage Vote Mechanism}

\author{
 \textbf{Kaixuan Du\textsuperscript{1}},
 \textbf{Meng Cao\textsuperscript{1}},
 \textbf{Hang Zhang\textsuperscript{1}},
 \textbf{Yukun Wang\textsuperscript{1}},\\
 \textbf{Xiangzhou Huang},
 \textbf{Ni Li\textsuperscript{1}\thanks{Corresponding author.}}
\\
 \textsuperscript{1}School of Automation Science and Electrical Engineering, Beihang University
\\
    \texttt{\{dukaixuan, lini\}@buaa.edu.cn}
}

\begin{document}
\maketitle 
\begin{abstract}
Current label-free RLVR approaches for large language models (LLMs), such as TTRL and Self-reward, have demonstrated effectiveness in improving the performance of LLMs on complex reasoning tasks. However, these methods rely heavily on accurate pseudo-label estimation and converge on spurious yet popular answers, thereby trapping in a dominant mode and limiting further improvements. Building on this, we propose \textbf{D}ual \textbf{C}onsensus \textbf{R}einforcement \textbf{L}earning\footnote{Code is available at \url{https://github.com/v0yager33/DualConsensus}.} (DCRL), a novel self-supervised training method which is capable of generating more reliable learning signals through a two-stage consensus mechanism. The model initially acts as an \textit{anchor}, producing dominant responses; then it serves as an \textit{explorer}, generating diverse auxiliary signals via a temporary unlearning process. The final training target is derived from the harmonic mean of these two signal sets. Notably, the process operates entirely without external models or supervision. Across eight benchmarks and diverse domains, DCRL consistently improves Pass@1 over majority vote while yielding more stable training dynamics. These results demonstrate that DCRL establishes a scalable path toward stronger reasoning without labels.
\end{abstract}

\section{Introduction}
 Reinforcement Learning with Verifiable Rewards (RLVR) has emerged as an effective approach to boosting the performance of Large Language Models (LLMs), enabling superior reasoning capabilities through long chain-of-thought \cite{wei2023chainofthought} reasoning on various challenging benchmarks \cite{openai2024openaio1card, deepseek-ai2025deepseekr1, yang2025qwen3}. However, typically implemented via algorithms such as Group Relative Policy Optimization (GRPO) \cite{shao2024deepseekmath}, current RLVR approaches heavily rely on human-annotated datasets or, at a minimum, environments that provide verifiable ground-truth signals \cite{le2022coderl, wang2024mathshepherdverifyreinforcellms}. This reliance restricts its generalizability to fully unlabeled or distribution-shifted tasks where neither human annotations \cite{ziegler2020finetuning} nor executable environments are accessible. As LLMs approach or surpass human-level performance, they will inevitably operate in domains where even expert humans cannot provide definitive judgments or reliable evaluations, which motivates the exploration of training on unlabeled data.
 
\begin{figure}[t]
  \includegraphics[width=\columnwidth]{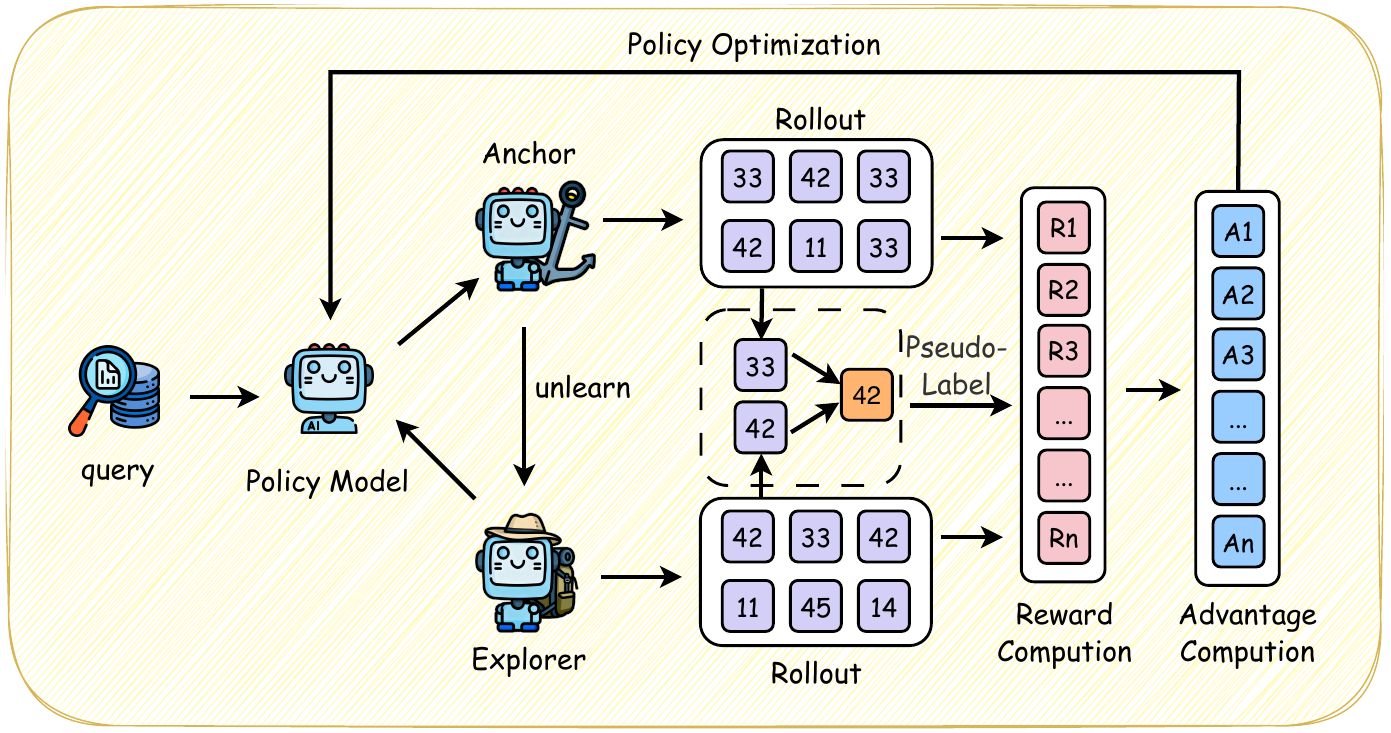}
  \caption{An overview of \textit{Dual Consensus Reinforcement Learning} (DCRL). Specifically, the policy model assumes two roles: (1) an \textit{anchor} that generates dominant and reliable responses; (2) an \textit{explorer} that produces diverse auxiliary signals through a temporary unlearning process.}
  \label{fig:method}
\end{figure}

 Recent studies \cite{shafayat2025can, zhao2025absoluteb} find that LLMs can achieve self-improvement without labeled data. One approach is determinism-based methods \cite{prabhudesai2025maximizing, zhang2025right}, which derive rewards from the confidence of a single policy along trajectories, thereby encouraging low-entropy and high-confidence predictions. These methods can achieve better performance by sharpening the model’s outputs, but it remains debatable whether they are truly effective for improving reasoning capabilities. Another approach is aggregation-based methods \cite{zhang2025corewarding,  yu2025restrain, wu2025spine}, which derive rewards from agreement across multiple samples, assuming that cross-sample consistency correlates with correctness. Nevertheless, these approaches still suffer from two critical limitations:
 \begin{itemize}
    \item \textbf{Spurious Reward Signals}: Models struggle to generate distinguishable reward signals, especially when tackling hard reasoning tasks; the answers derived from majority vote may themselves suffer from systematic biases \cite{zhao2025majority}. In the later stages of training, spurious majority outcomes can come to dominate, yet the correct solutions may instead lie within the minority rollouts.
    \item \textbf{Lack Exploration Capability}: By continuously rewarding consensus across diverse trajectories, the model’s output distribution becomes increasingly rigid and concentrated, resulting in a severe deficiency in exploration capability or even entropy collapse \cite{cui2025entropymechanismreinforcementlearning}. Consequently, the model tends to converge to a narrow set of suboptimal responses and exhibits degraded performance when confronted with out-of-domain (OOD) tasks.
\end{itemize}

 In this paper, we propose \textit{Dual Consensus}—a novel framework for Unsupervised Reinforcement Learning with Verifiable Rewards (URLVR) driven by a multi-stage vote mechanism. Our core insight stems from the following intuition: valid reasoning trajectories should not only converge to the dominant mode but also exhibit enhanced robustness when the distribution of the model is artificially flattened. 
 
 Instead of naively adopting the fragile majority vote as the pseudo-label, we decompose the rollout process into two stages: \textit{anchor} and \textit{explorer}. The \textit{anchor} stage involves normal rollouts where the model generates responses under its current policy, capturing the dominant reasoning mode. Subsequently, in the \textit{explorer} stage, we introduce a temporary unlearning process to flatten the distribution and enhance exploration, thereby encouraging the generation of diverse auxiliary responses that deviate from the dominant mode. After obtaining the two signal sets, we compute the harmonic mean of their consensus scores to determine the final reward signal. This harmonic mean balances the reliability of the dominant mode (from the anchor stage) and the diversity of potential valid trajectories (from the explorer stage), effectively mitigating the adverse impact of majority vote when it converges to spurious answers.

 To validate our approach, we demonstrate the effectiveness of \textit{Dual Consensus} through extensive experiments. We first train models on the large-scale DAPO-14K-Math dataset and evaluate them on multiple established benchmarks. Additionally, we apply test-time adaptation on five distinct datasets to further assess our method. In summary, our key contributions are as follows:
 
 \begin{itemize}
    \item  A novel URLVR method, \textit{Dual Consensus}, which utilizes the intrinsic robustness of the model to guide the model to evolve with policy optimization methods such as GRPO. Notably, the framework is entirely free of external models and enables continuous self-improvement without supervision.
    \item A pseudo-label selection mechanism and reward design that exploits the intrinsic robustness of the model itself to generate more reliable reward signals, mitigating the toxicity of majority vote when it fails.
    \item We empirically verify the general effectiveness of \textit{Dual Consensus} in boosting LLMs’ reasoning performance via comprehensive experiments, and additionally present systematic ablation studies and in-depth further analyses.
\end{itemize}

\section{Methodology}
\label{sec:methodology}
In this section, we present the details of Dual Consensus. It mitigates spurious majority bias by generating diverse signals via Unlearn Then Explore, identifying more accurate labels through Harmonic Election, and stabilizing updates with Adaptive Sampling.

Our framework employs Grouped Relative Policy Optimization (GRPO) \cite{shao2024deepseekmath} as its foundational RL algorithm—it stabilizes training by normalizing advantage estimates across multiple rollouts of the same prompt.

For a given input prompt $x$ (paired with its ground-truth label $y$), GRPO first samples $n$ rollouts $\{y_i\}_{i=1}^n$ from the current policy. For each rollout $y_i$, GRPO computes a reward $r_i = R(y_i, y \mid x)$, then derives the group-normalized advantage $\hat{A}_i$:
\begin{equation}
\hat{A}_i = \frac{r_i - \bar{r}}{\sigma_r}
\end{equation}
\noindent where $\bar{r} = \frac{1}{n}\sum_{k=1}^n r_k$ denotes the mean reward of the rollout group, and $\sigma_r$ is the standard deviation of the group rewards. The GRPO objective optimizes the target policy $\pi_\theta$ by maximizing the clipped normalized advantage:
\begin{multline}
\mathcal{L}_{\text{GRPO}}(x, y; \theta) = \\
\frac{1}{n} \sum_{i=1}^n \min\bigl( \rho_i(\theta) \hat{A}_i,\; \text{clip}\bigl(\rho_i(\theta), 1-\epsilon, 1+\epsilon\bigr) \hat{A}_i \bigr) \\
- \beta \, \mathbb{D}_{\mathrm{KL}}\bigl[ \pi_\theta(\cdot \mid x) \parallel \pi_{\text{ref}}(\cdot \mid x) \bigr]
\end{multline}
where $\rho_i(\theta) = \pi_\theta(y_i \mid x)/\pi_{\text{old}}(y_i \mid x)$ is the importance sampling ratio.

\subsection{Unlearn Then Explore}

Following the GRPO framework, we first initialize an \textit{\textbf{anchor}} model (parameterized by $\theta'$) by cloning the current policy model (parameterized by $\theta$), i.e., $\theta' \leftarrow \theta$. We then strategically apply an unlearning strategy \cite{liu2024rethinkingmachineunlearninglarge} to transform this \textit{anchor} into an \textit{\textbf{explorer}}. Unlike EEPO \cite{chen2025eepo}, which simply employs unlearning as a supplementary technique to enhance exploration, our approach leverages unlearning as a core methodology to actively search for correct answers. 

To implement this unlearning strategy, we firstly introduce the standard \textit{negative log-likelihood} (NLL) loss that is defined as:
\begin{equation}
\mathcal{L}_{\text{NLL}} = -\log \pi_{\text{anchor}}(y_{i,t} \mid x, y_{i,<t})
\end{equation}
This loss function imposes heavier penalties on predictions with low probability. Conversely, we take a complementary loss function that inverts the penalty pattern of the NLL loss. 

To ensure numerical stability when $\pi_{\text{anchor}}(y_{i,t} \mid x, y_{i,<t}) \to 1$ (which would make $1 - \pi_{\text{anchor}}(y_{i,t} \mid x, y_{i,<t})$ approach zero and cause numerical overflow), we first perform a clipping operation on the prediction probability from the \textit{anchor} model:
\begin{equation}
p_{\text{clip}} = \mathrm{clip}\bigl(\pi_{\text{anchor}}(y_{i,t} \mid x, y_{i,<t}),\,\epsilon,\,1-\epsilon\bigr)
\end{equation}
where $\epsilon$ is a small positive constant for numerical stability. This clipping operation both avoids the term $1 - \pi_{\text{anchor}}(y_{i,t} \mid x, y_{i,<t})$ from becoming excessively small and eliminates unnecessary penalties for tokens with extremely low probabilities that are irrelevant to meaningful exploration.

Based on the clipped probability $p_{\text{clip}}$, we define the stabilized unlearning loss as follows:
\begin{equation}
\mathcal{L}_{\text{unlearn}} = -\log\bigl(1 - p_{\text{clip}}\bigr)
\label{eq:unlearn_loss}
\end{equation}
This loss function achieves the targeted penalty characteristic: high $p_{\text{clip}}$ tokens lead to a large negative $\log(1 - p_{\text{clip}})$, and minimizing the loss enforces a sharp reduction in their probabilities with strong penalties.
\begin{figure}[t]
  \centering
  \includegraphics[width=\columnwidth]{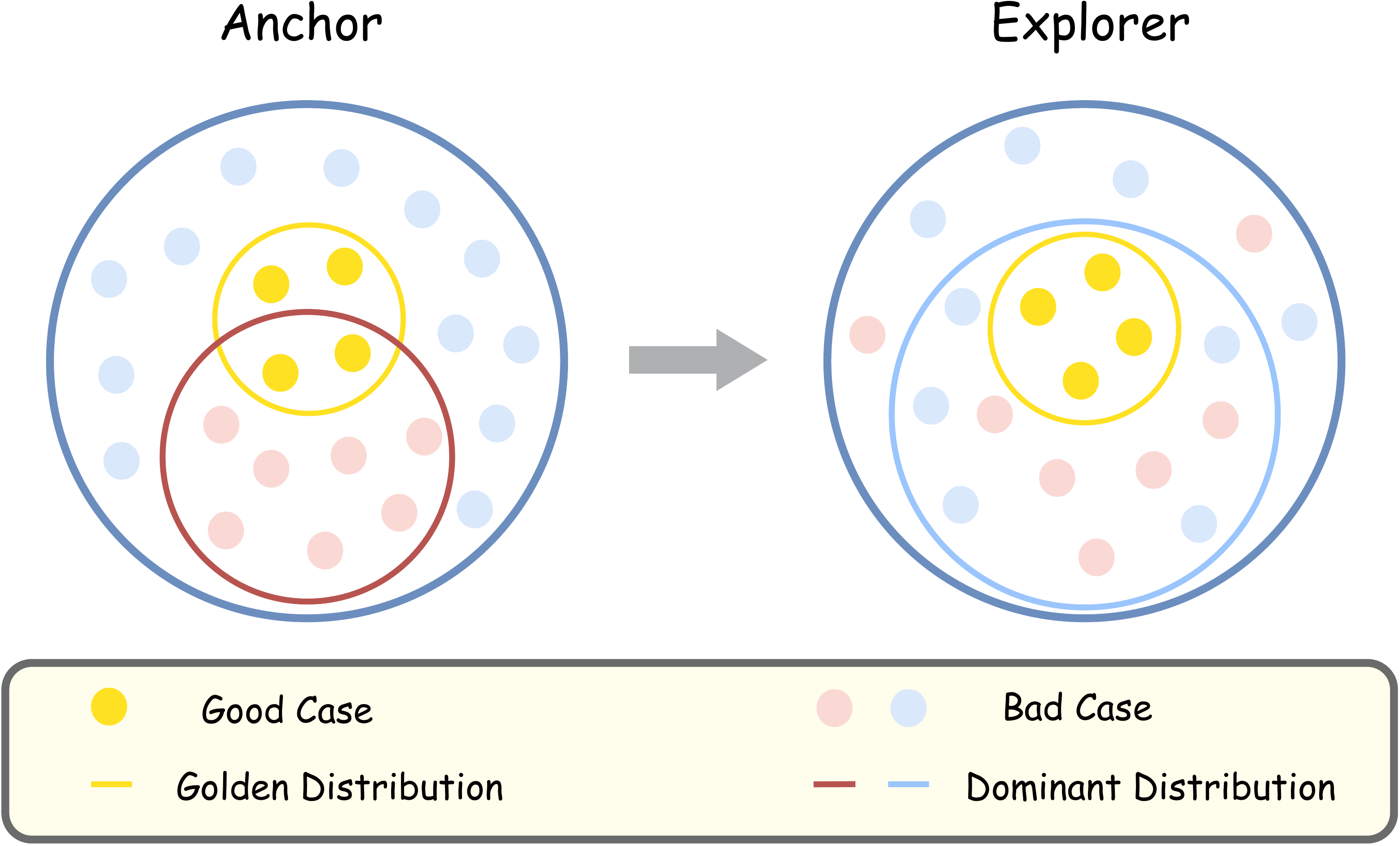}
  \caption{Output distributions of the \textit{Anchor} and \textit{Explorer} models. The \textit{Explorer} model, after the unlearning process, generates a more diverse distribution.}
  \label{fig:distribution}
\end{figure}

To implement this unlearning process in practice, we apply a single gradient descent step to the \textit{anchor} model using the unlearn loss $\mathcal{L}_{\text{unlearn}}$, which transforms it into an \textit{explorer} model:
\begin{equation}
\theta' \leftarrow \theta' - \eta \nabla_{\theta'} \mathcal{L}_{\text{unlearn}}(\theta')
\end{equation}
where $\eta$ denotes the learning rate for this unlearning step. Critically, this update is \textit{temporary}—it is confined exclusively to the \textit{anchor} model within the current iteration, and the parameters of the original policy model $\theta$ remain unchanged.

The gradient update direction $-\eta \nabla_{\theta'} \mathcal{L}_{\text{unlearn}}$ suppresses high-confidence tokens from the anchor model. The derivative of $\mathcal{L}_{\text{unlearn}} = -\log(1-p_{\text{clip}})$ assigns larger gradient magnitudes to tokens with higher $p_{\text{clip}}$ values, and gradient descent directly drives the reduction of their probabilities to achieve the unlearning effect.

As illustrated in Fig.~\ref{fig:distribution}, the explorer model modulates the anchor’s probability distribution, expanding the coverage of generated responses to include trajectories that deviate from the dominant mode, which provides the necessary diverse reasoning signals for subsequent consensus aggregation.
\subsection{Harmonic Election}
\begin{figure*}[t]
  \centering
  \includegraphics[width=0.95\textwidth]{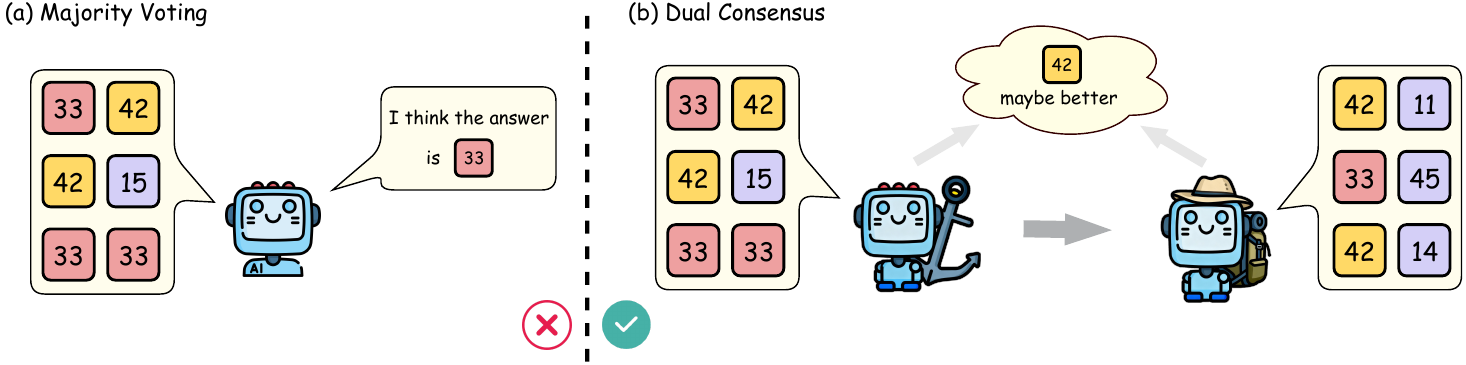}
  \caption{Comparison between \textit{Majority vote} and \textit{Dual Consensus}: Majority vote tends to fall into spurious consensus by over-relying on dominant but potentially incorrect response modes, while \textit{Dual Consensus} mitigates this issue by converting the anchor model (which captures dominant reasoning patterns) into an explorer model via temporary unlearning. This transformation enables the framework to explore diverse alternative response modes, thereby balancing the reliability of current dominant patterns and the diversity of potential valid alternatives, and ultimately achieving more accurate answer selection.}
  \label{fig:contrast}
\end{figure*}
Instead of relying on majority vote, we employ the harmonic mean to establish consensus between the anchor and explorer models, which effectively balances exploration and exploitation (shown in Fig \ref{fig:contrast}). This consensus process proceeds in three sequential steps.

First, we perform anchor rollout: drawing $G$ trajectories from the anchor policy $\pi_{\text{anchor}}$, denoted as:
\begin{equation}
O_0 = \{o_1, o_2, \dots, o_G\}.
\label{eq:trajectories_anchor}
\end{equation}
Then we apply temporary unlearning to $\pi_{\text{anchor}}$ by minimizing the unlearning loss (Eq.~\ref{eq:unlearn_loss}), thereby suppressing its dominant generation patterns and converting it into the explorer model $\pi_{\text{explorer}}$.

We subsequently conduct explorer rollout: generating another $G$ trajectories from $\pi_{\text{explorer}}$, forming the set:
\begin{equation}
O_1 = \{o_1', o_2', \dots, o_G'\}.
\label{eq:trajectories_explorer}
\end{equation}
Let $\mathcal{A}$ be the set of all candidate answers. For each $a \in \mathcal{A}$, we compute its empirical occurrence probabilities in $O_0$ and $O_1$:
\begin{align}
p_0(a) &= \frac{1}{G}\sum_{i=1}^G \mathbb{I}\bigl(o_i \mapsto a\bigr), \\
p_1(a) &= \frac{1}{G}\sum_{i=1}^G \mathbb{I}\bigl(o_i' \mapsto a\bigr),
\end{align}
where $\mathbb{I}(\cdot)$ is the indicator function.

The consensus pseudo-label $y^*$ is then selected as the answer that maximizes the harmonic mean of $p_0(a)$ and $p_1(a)$:
\begin{equation}
y^* = \mathop{\arg\max}_{a\in\mathcal{A}} \frac{2 p_0(a) p_1(a)}{p_0(a)+p_1(a)}.
\label{eq:harmonic_consensus}
\end{equation}

We note \textit{Self-Harmony} \cite{wang2025selfharmonylearningharmonizeselfsupervision} also uses the harmonic mean, yet our method only processes the model’s distribution for the same input, thus avoiding Self-Harmony’s semantic inconsistency in question rephrasing—an frequently recurring issue with catastrophic consequences.

During reward computation, a full reward is assigned to trajectories generating the consensus pseudo-label $y^*$. Trajectories consistent with the majority result of the anchor are assigned a modest reserved reward to avoid negative advantages in subsequent estimation, since such trajectories are still deemed promising relative to other counterparts. Formally, the reward for trajectory $i$ is defined as:
\begin{equation}
r_i =
\begin{cases}
1 & \text{if } y_i(o) = y^{*}, \\
0.5 & \text{if } y_i(o) = \hat{y}_{\text{anchor}}, \\
0 & \text{otherwise},
\end{cases}
\label{eq:reward}
\end{equation}
where $\hat{y}_{\text{anchor}} = \mathop{\arg\max}_{a\in\mathcal{A}} \sum_{o_i\in O_0}\mathbb{I}\bigl(o_i \mapsto a\bigr)$ denotes the majority answer from the anchor trajectories.

\subsection{Adaptive Sampling}
Unlike the static sampling strategy in majority vote~\cite{zuo2025ttrl}, we introduce the \textit{consensus rate} as a signal to adaptively regulate the contribution of \textit{anchor} and \textit{explorer} rollouts during policy updates.

We formally define the consensus rate $\rho_t$ at step $t$ as the proportion of anchor-sampled trajectories whose generated answers are consistent with the majority-voted result $\hat{y}_{\text{anchor}}$ of the anchor stage, which is formulated as:
\begin{equation}
\rho_t = \frac{1}{|O_0|} \sum_{o \in O_0} \mathbb{I}\bigl(y(o) = \hat{y}_{\text{anchor}}\bigr)
\label{eq:consensus_rate}
\end{equation}
where $y(o)$ denotes the answer extracted from trajectory $o$, and $\mathbb{I}(\cdot)$ is the indicator function.

To capture long-term consistency and mitigate step-wise noise, we maintain a sliding window of the consensus rate over the most recent $K$ steps and compute its mean:
\begin{equation}
\bar{\rho}_t = \frac{1}{K} \sum_{k=1}^{K} \rho_{t - k + 1}.
\end{equation}
A high $\bar{\rho}_t$ indicates that the policy model consistently converges to the same answer, reflecting strong model certainty and deterministic behavior; conversely, a low $\bar{\rho}_t$ suggests ongoing exploration.

Based on $\bar{\rho}_t$, we design a dynamic sampling selection rule with threshold $1/2$:
\begin{itemize}
    \item When $\bar{\rho}_t \leq \tfrac{1}{2}$, only anchor trajectories $O_0$ are used for policy updates. Explorer trajectories $O_1$ still participate in consensus formation via harmonic vote but are excluded from gradient computation. This prevents premature incorporation of noisy exploratory signals while preserving rare answers that align with the pseudo-label.
    \item When $\bar{\rho}_t > \tfrac{1}{2}$, both $O_0$ and $O_1$ are included in training. This enables the policy to leverage reliable anchor behaviors while actively integrating diverse, high-quality explorations.
\end{itemize}
The effective rollout set for policy update is defined as:
\begin{equation}
\mathcal{O}_{\text{train}} =
\begin{cases}
O_0 & \text{if } \bar{\rho}_t \leq \tfrac{1}{2}, \\
O_0 \cup O_1 & \text{if } \bar{\rho}_t > \tfrac{1}{2}.
\end{cases}
\label{eq:dynamic_sampling}
\end{equation}
Only trajectories in $\mathcal{O}_{\text{train}}$ contribute to the policy gradient, ensuring a smooth transition from exploitation-dominant to balanced exploration-exploitation learning.

\begin{algorithm}[ht]
\caption{Dual Consensus: An Unsupervised RLVR Algorithm}
\begin{algorithmic}[1]
\State \textbf{Initialize:} policy $\theta^0$; learning rates $\eta_{\text{GRPO}}, \eta_{\text{u}}$; group size $G$; iterations $T$
\For{$t = 0$ to $T - 1$}
    \State Sample query $x \sim \mathcal{D}$
    \State Sample $G$ trajectories $O_{\text{anchor}} \sim \pi_{\theta^t}(\cdot \mid x)$, compute consensus rate $\rho_t$ and $\bar{\rho}_t$
    \State $\theta_{\text{e}} \leftarrow \theta^t - \eta_{\text{u}} \nabla_{\theta^t} \mathcal{L}_{\text{u}}(O_{\text{anchor}})$
    \State Sample $G$ trajectories $O_{\text{explorer}} \sim \pi_{\theta_{\text{e}}}(\cdot \mid x)$
    \State $S(a) = \frac{2p_0(a)p_1(a)}{p_0(a)+p_1(a)}$ for each answer $a$ in $O_{\text{anchor}} \cup O_{\text{explorer}}$
    \State Pseudo-label $y^* = \arg\max_{a} S(a)$
    \If{$\rho_t > 1/2$}
        \State $O \leftarrow O_{\text{anchor}} \cup O_{\text{explorer}}$
    \Else
        \State $O \leftarrow O_{\text{anchor}}$
    \EndIf
    \State Compute rewards and advantages for $O$
    \State $\theta^{t+1} \leftarrow \theta^t + \eta_{\text{GRPO}} \nabla_\theta J_{\text{GRPO}}(\theta^t; O)$
\EndFor
\end{algorithmic}
\end{algorithm}
The overall workflow of the proposed Dual Consensus algorithm is summarized in Algorithm 1.

\begin{table*}[h]
    \centering
    \small
    \setlength{\tabcolsep}{2mm} 
    \begin{tabular}{c|cccccccc|c}
        \toprule[0.8pt]
        \multirow{2}{*}{\textbf{Methods}} & \multicolumn{6}{c}{\textbf{Math}} & \multicolumn{2}{c|}{\textbf{Multi-Task}} & \multirow{2}{*}{\textbf{Average}} \\
        \cline{2-7} \cline{8-9}
        & \textbf{Math} & \textbf{GSM8K} & \textbf{AIME24} & \textbf{Minerva.} & \textbf{AMC} & \textbf{Olympiad.} & \textbf{MMLU.} & \textbf{GPQA.} &  \\
        \midrule[0.5pt]
        \multicolumn{10}{c}{\textit{\textbf{Llama3.2-3B-Instruct}}} \\
        \midrule[0.3pt]
        Vanilla & 42.4 & 76.1 & 4.5 & 11.7 & 20.6 & 14.6 & 26.4 & 22.3 & 27.3 \\
        GRPO & 49.2 & 79.7 & 13.5 & 14.9 & 23.2 & 15.3 & 32.8 & 23.8 & 31.5 \\
        \midrule[0.3pt]
        RENT & 45.4 & 78.5 & 9.4 & 11.2 & 20.9 & 15.2 & 30.3 & 22.7 & 29.2 \\
        TTRL & 44.6 & 71.8 & 10.2 & 12.4 & 21.5 & 14.6 & 25.9 & 21.4 & 27.8 \\
        Co-Rewarding-I & 45.3 & 77.4 & \textbf{11.8} & 15.1 & 22.2 & 14.8 & \textbf{33.9} & 24.3 & 30.6 \\
        Co-Rewarding-II & 47.6 & 76.9 & 11.6 & 13.7 & \textbf{23.3} & 15.3 & 24.9 & 21.4 & 29.3 \\
        \textbf{\textit{DCRL(ours)}} & \textbf{47.8} & \textbf{79.1} & \textbf{11.8} & \textbf{15.8} & \textbf{23.3} & \textbf{15.4} & \textbf{33.9} & \textbf{24.7} & \textbf{31.4} \\
        \midrule[0.5pt]
        \multicolumn{10}{c}{\textit{\textbf{Qwen3-4B-Base}}} \\
        \midrule[0.3pt]
        Vanilla & 47.4 & 87.0 & 9.3 & 24.5 & 38.3 & 34.9 & 48.6 & 30.7 & 40.0 \\
        GRPO & 76.8 & 91.8 & 13.1 & 32.9 & 45.2 & 36.1 & 52.2 & 34.8 & 47.8 \\
        \midrule[0.3pt]
        RENT & 72.1 & 85.0 & 10.2 & 23.7 & 44.3 & 36.6 & 48.6 & 32.2 & 44.0 \\
        TTRL & 74.4 & 91.3 & 11.5 & 29.0 & 44.7 & 37.0 & 51.2 & 32.8 & 46.4 \\
        Co-Rewarding-I & \textbf{74.8} & 91.6 & 11.8 & 31.1 & 43.1 & 36.9 & 52.2 & 32.9 & 46.7 \\
        Co-Rewarding-II & 74.5 & 91.7 & 11.6 & 30.8 & 44.5 & 37.1 & \textbf{52.6} & 33.3 & 47.0 \\
        \textbf{\textit{DCRL(ours)}} & \textbf{74.8} & \textbf{91.7} & \textbf{12.4} & \textbf{31.2} & \textbf{45.6} & \textbf{37.2} & \textbf{52.6} & \textbf{34.3} & \textbf{47.4} \\
        \midrule[0.5pt]
        \multicolumn{10}{c}{\textit{\textbf{Qwen3-8B-Base}}} \\
        \midrule[0.3pt]
        Vanilla & 61.8 & 88.3 & 12.5 & 25.0 & 48.9 & 40.1 & 50.6 & 32.3 & 44.9 \\
        GRPO & 82.0 & 93.8 & 20.4 & 34.1 & 57.0 & 45.8 & 57.5 & 38.1 & 53.5 \\
        \midrule[0.3pt]
        RENT & 75.3 & 86.5 & 13.1 & 25.4 & 49.7 & 40.3 & 52.5 & 34.5 & 47.1 \\
        TTRL & 78.3 & 92.9 & 14.4 & \textbf{32.7} & 51.2 & \textbf{40.9} & 55.2 & 36.5 & 50.2 \\
        Co-Rewarding-I & 78.2 & 92.7 & 12.2 & 32.0 & 51.2 & 40.7 & \textbf{56.7} & \textbf{37.9} & 50.2 \\
        Co-Rewarding-II & 77.2 & 92.9 & 12.2 & 32.3 & 51.8 & 40.2 & 56.1 & 37.6 & 50.0 \\
        \textbf{\textit{DCRL(ours)}} & \textbf{79.2} & \textbf{93.3} & \textbf{14.7} & \textbf{32.7} & \textbf{51.9} & \textbf{40.9} & \textbf{56.7} & \textbf{37.9} & \textbf{50.9} \\
        \bottomrule[0.8pt]
    \end{tabular}
    \caption{\textbf{Main Results (\%) of DCRL Trained on DAPO-Math-14k: DCRL Outperforms All Label-Free Baselines.} The best results are highlighted in \textbf{bold}. DCRL exceeds all unsupervised methods and nearly matches GRPO with gold labels.}
    \label{tab:main_results}
\end{table*}

\section{Experiments}
In this section, we first introduce the experimental setup, then discuss the overall effectiveness of our method, and finally present the results of our ablation studies.
\subsection{Setups}
To comprehensively evaluate the effectiveness of DCRL, our experiments are conducted under two distinct training paradigms:
\begin{itemize}[leftmargin=12pt,itemsep=0pt,topsep=0pt]
    \item \textbf{Large-Scale Unsupervised Learning:} Directly applying DCRL to train models from scratch on a large, unlabeled dataset.
    \item \textbf{Test-Time Adaptation (TTA):} Using DCRL to adapt a pre-trained model to new, unseen benchmarks via constant unsupervised training.
\end{itemize}

\paragraph{Models:} Our target models include Llama3.2-3B-Instruct \cite{grattafiori2024llamaa}, Qwen3-4B-Base, and Qwen3-8B-Base \cite{yang2025qwen3}. Additionally, for the TTA paradigm, we also conduct experiments on Qwen2.5-Math-1.5B \cite{yang2024qwen25math}.

\paragraph{Implementation:} Our primary training dataset is DAPO-Math-14k, a processed version of DAPO-Math-17k \cite{yu2025dapoa}, which is refined by deduplicating prompts and standardizing the formatting of both prompts and reference answers. We train each model on this dataset for two epochs to mitigate overfitting. All experiments are based on the VeRL framework \cite{sheng2025hybridflow}. Implementation details are reported in Appendix \ref{sec:implementation details}.

\paragraph{Benchmarks:} Our benchmark suite comprises eight challenging datasets, including six math-specific: (1) MATH-500 \cite{hendrycks2021measuring}, (2) GSM8K \cite{cobbe2021training}, (3) AIME24 \cite{yang2025qwen3}, (4) Minerva-math \cite{lewkowycz2022solvingquantitativereasoningproblems}, (5) AMC \cite{zuo2025ttrl}, (6) OlympiadBench \cite{he-etal-2024-olympiadbench};  and two multi-task benchmarks: (1) MMLU-Pro \cite{NEURIPS2024_ad236edc}, (2) GPQA-Diamond \cite{rein2024gpqa}. For the TTA experiments, we evaluate on subsets of MATH-500, AIME24, AMC, and GPQA-Diamond.
\begin{figure}[t!]
  \centering
  \begin{subfigure}{\linewidth}
    \includegraphics[width=\linewidth]{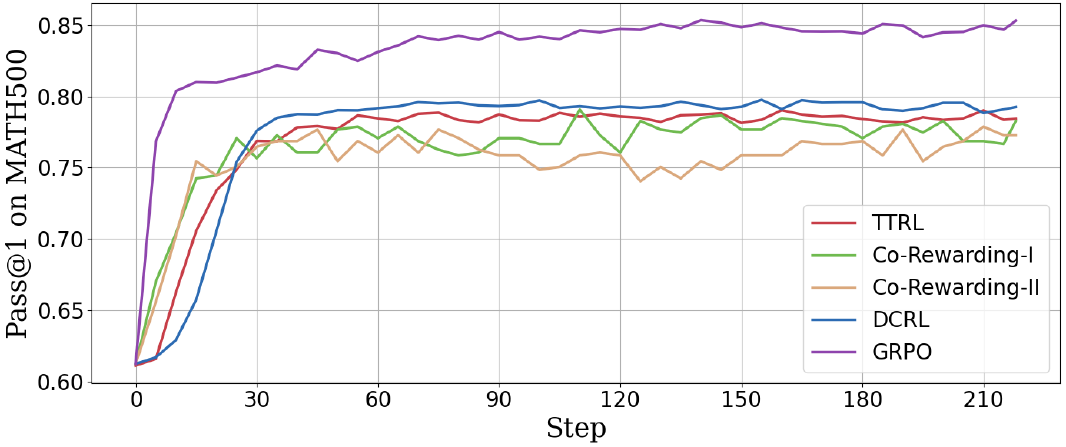}
    \caption{Accuracy curve on the MATH500 benchmark.}
    \label{subfig:math_acc_curve}
  \end{subfigure}
  \begin{subfigure}{\linewidth}
    \includegraphics[width=\linewidth]{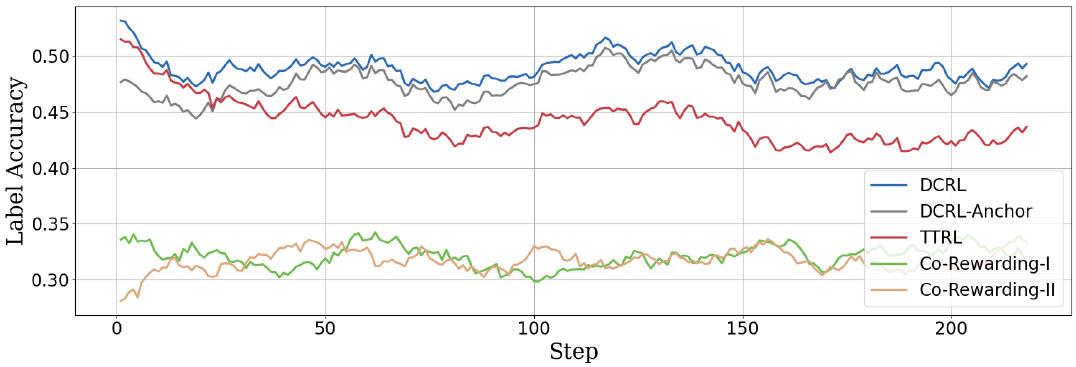}
    \caption{Label accuracy curve (smoothed).}
    \label{subfig:label_acc_curve}
  \end{subfigure}
  \caption{Training Dynamics of Dual Consensus on Qwen3-8B-Base. DCRL-Anchor in Fig. \ref{subfig:label_acc_curve} refers to the majority vote of the anchor model in DCRL.}
  \label{fig:training_dynamics}
\end{figure}
\paragraph{Baselines:} We compare DCRL against four unsupervised RLVR methods, including one determinism-based approach \textbf{RENT} \cite{prabhudesai2025maximizing}, and three aggregation-based approaches \textbf{TTRL} \cite{zuo2025ttrl}, \textbf{Co-Rewarding-I}, and \textbf{Co-Rewarding-II} \cite{zhang2025corewarding}. Specifically, for Co-Rewarding-I, we adopt a dataset rephrased by Qwen3-32B for all related experiments.

\paragraph{Metrics:} We use the Pass@1 metric. For each question, we sample 16 predictions using a temperature of 0.6 and a top-p value of 0.95. The final reported score is the average Pass@1 accuracy across these 16 independent seeds.
\subsection{Results}

\subsubsection{Main Performance of Dual Consensus}
\label{sec:main performance}
Table \ref{tab:main_results} presents the main results of DCRL trained on DAPO-Math-14k across eight challenging reasoning benchmarks. Our method consistently outperforms all label-free baselines—including RENT, TTRL, and both variants of Co-Rewarding—across different model scales and task domains. Notably, on the Qwen3-8B-Base model, DCRL achieves 79.2\% on MATH-500, surpassing the strongest baseline TTRL at 78.3\% by 0.9\%, and improves AIME24 from 14.4\% to 14.7\%, demonstrating its effectiveness on extremely hard competition-level problems. On multi-task benchmarks, DCRL matches or exceeds Co-Rewarding-I on MMLU-Pro and GPQA-Diamond, confirming its generalizability beyond pure math reasoning.

Remarkably, despite being fully unsupervised and using no ground-truth labels, DCRL achieves performance on par with, and occasionally exceeds, the supervised GRPO baseline. On Llama3.2-3B-Instruct, Qwen3-4B-Base, and Qwen3-8B-Base, it yields average gains of +7.5\%, +4.0\%, and +6.1\% on MMLU-Pro respectively.
\begin{figure}[h]
    \begin{minipage}{0.48\textwidth}
        \centering
        \includegraphics[width=\textwidth]{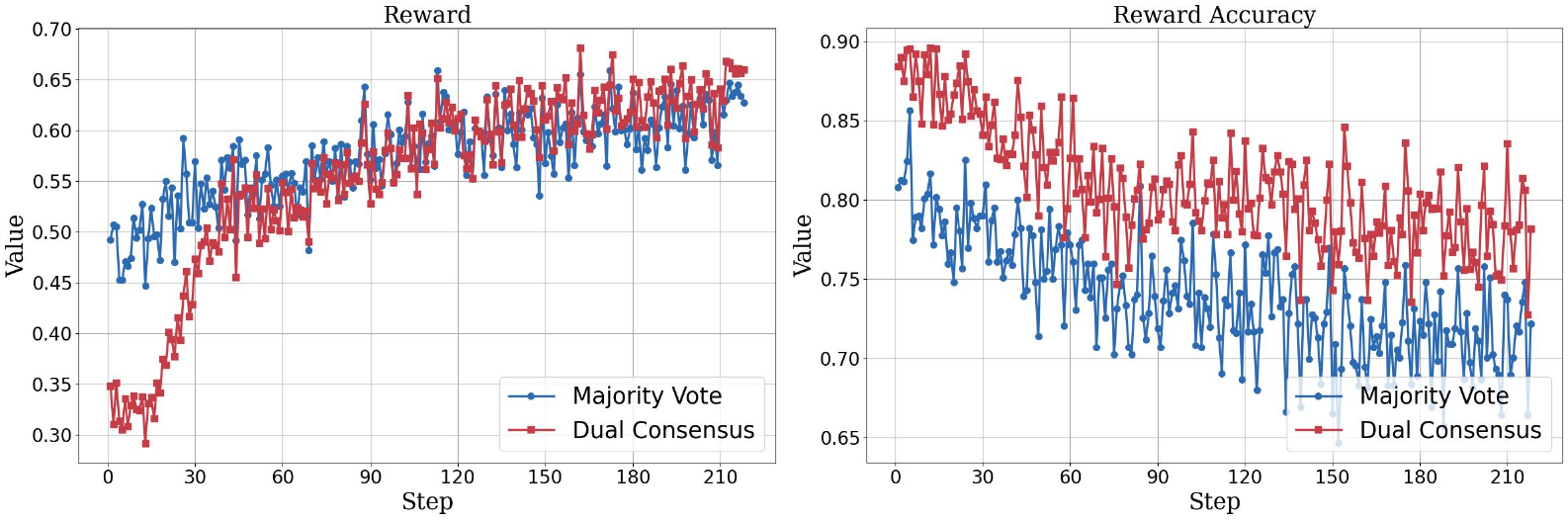}
    \end{minipage}
    \caption{Comparison of reward signal between Majority Vote and Dual Consensus.}
    \label{fig:reward_comparison}
\end{figure}

\paragraph{Compared with Majority Vote:} We adopt TTRL \cite{zuo2025ttrl} as a representative baseline that relies on standard majority vote for pseudo-label estimation. Fig \ref{fig:training_dynamics} and Fig \ref{fig:reward_comparison} depict the training dynamics on Qwen3-8B-Base: our method can select low-consistency answers (especially early in training) and achieves higher ground-truth reward accuracy. This demonstrates that Dual Consensus produces more reliable supervision signals than naive majority vote.
\begin{table}[h]
    \centering
    \tiny
    \setlength{\tabcolsep}{2.2mm}
    \renewcommand{\arraystretch}{1.05}
    \begin{tabular}{l|cccc|c}
        \toprule[0.8pt]
        \multirow{2}{*}{\textbf{Methods}} & \multicolumn{4}{c|}{\textbf{Datasets}} & \multirow{2}{*}{\textbf{Average}} \\
        \cmidrule(lr){2-5}
        & \textbf{MATH500} & \textbf{AIME24} & \textbf{AMC} & \textbf{GPQA} & \\
        \midrule[0.5pt]
        \multicolumn{6}{c}{\textit{\textbf{Qwen2.5-Math-1.5B}}} \\
        \midrule[0.3pt]
        \addlinespace[0.3em]
        Vanilla & 30.6 & 5.6 & 23.4 & 15.4 & 18.7 \\
        \midrule[0.3pt]
        TTRL & 72.9 & 17.0 & 45.3 & 22.2 & 39.3 \\
        \textbf{\textit{DCRL(ours)}} & 74.5 & 17.7 & 46.6 & 22.8 & 40.4 \\
        \midrule[0.3pt]
        $\Delta$(-TTRL) & $\uparrow$2.2\% & $\uparrow$4.1\% & $\uparrow$2.8\% & $\uparrow$2.7\% & $\uparrow$2.7\% \\
        \midrule[0.5pt]
        \multicolumn{6}{c}{\textit{\textbf{Llama3.2-3B-Instruct}}} \\
        \midrule[0.3pt]
        \addlinespace[0.3em]
        Vanilla & 42.4 & 4.5 & 20.6 & 22.3 & 22.4 \\
        \midrule[0.3pt]
        TTRL & 59.6 & 10.0 & 26.5 & 28.9 & 31.2 \\
        \textbf{\textit{DCRL(ours)}} & 59.8 & 13.3 & 32.3 & 32.6 & 34.5 \\
        \midrule[0.3pt]
        $\Delta$(-TTRL) & $\uparrow$0.3\% & $\uparrow$33.0\% & $\uparrow$21.8\% & $\uparrow$12.8\% & $\uparrow$10.5\% \\
        \midrule[0.5pt]
        \multicolumn{6}{c}{\textit{\textbf{Qwen3-4B-Base}}} \\
        \midrule[0.3pt]
        \addlinespace[0.3em]
        Vanilla & 47.4 & 9.3 & 38.3 & 30.7 & 31.4 \\
        \midrule[0.3pt]
        TTRL & 82.6 & 17.2 & 56.5 & 35.3 & 47.9 \\
        \textbf{\textit{DCRL(ours)}} & 83.4 & 20.6 & 56.5 & 35.6 & 49.0 \\
        \midrule[0.3pt]
        $\Delta$(-TTRL) & $\uparrow$0.9\% & $\uparrow$19.7\% & $\uparrow$0.0\% & $\uparrow$0.8\% & $\uparrow$2.2\% \\
        \midrule[0.5pt]
        \multicolumn{6}{c}{\textit{\textbf{Qwen3-8B-Base}}} \\
        \midrule[0.3pt]
        \addlinespace[0.3em]
        Vanilla & 61.8 & 12.5 & 48.9 & 32.3 & 38.8 \\
        \midrule[0.3pt]
        TTRL & 85.7 & 19.8 & 59.8 & 43.1 & 52.1 \\
        \textbf{\textit{DCRL(ours)}} & 86.4 & 22.8 & 61.0 & 44.5 & 53.6 \\
        \midrule[0.3pt]
        $\Delta$(-TTRL) & $\uparrow$0.8\% & $\uparrow$15.1\% & $\uparrow$2.0\% & $\uparrow$3.2\% & $\uparrow$2.8\% \\
        \bottomrule[0.8pt]
    \end{tabular}
    \caption{\textbf{Results (\%) of Test-Time Adaptation Trained on Different Datasets: DCRL Consistently Outperforms the TTRL Baseline.} All results are evaluated under the same experimental settings and reported as the average pass@1 over 16 independent seeds.}
    \label{tab:Test-time Adaption}
\end{table}

\subsubsection {Test-time Adaption}
\begin{figure}[h]
    \begin{minipage}{0.48\textwidth}
        \centering
        \includegraphics[width=\textwidth]{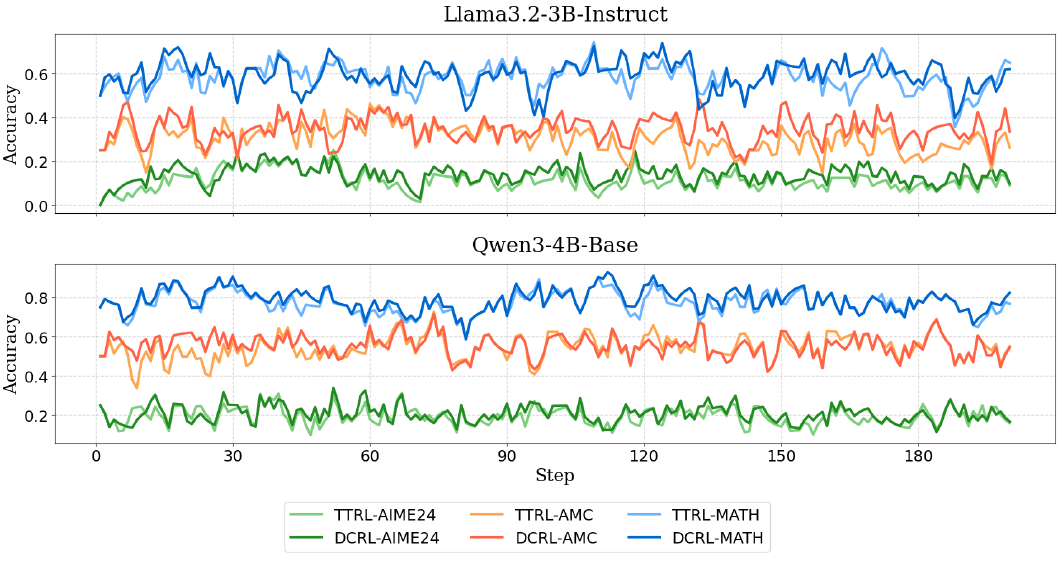}
    \end{minipage}
    \caption{Label accuracy curves (smoothed) of test-time adaptation for DCRL and TTRL across different tasks.}
    \label{fig:TTA_label_accuracy}
\end{figure}
Test-time Adaptation (TTA) serves as a critical validation scenario for unsupervised RLVR methods, as it directly evaluates the ability to escape spurious majority bias and generalize to unseen reasoning tasks without labeled data. The performance of TTA is presented in Table \ref{tab:Test-time Adaption}. We did not include methods such as \textit{RESTRAIN} \cite{yu2025restrain} and \textit{Self-Harmony} \cite{liu2025ettrl} as baselines because their official code has not been released.

Although TTA still enables generalization to other unseen scenarios with limited data \cite{shafayat2025can, zuo2025ttrl}, a defining characteristic of TTA is its immunity to overfitting concerns. In this setting, the model’s ability to a priority avoid spurious reward signals becomes critically important. As demonstrated in Table \ref{tab:Test-time Adaption}, our DCRL consistently outperforms TTRL across all evaluated tasks, validating the efficacy of our dual consensus mechanism in suppressing misleading signals.

\subsection{Ablation Studies}
\begin{table*}[h]
    \centering
    \small
    \setlength{\tabcolsep}{4.2mm}
    \renewcommand{\arraystretch}{1.05}
    \begin{tabular}{l|ccccc|c}
        \toprule[0.8pt]
        \textbf{Method}          & \textbf{MATH} & \textbf{GSM8K} & \textbf{AIME24} & \textbf{MMLU.} & \textbf{GPQA.} & \textbf{Average} \\
        \midrule
        \textbf{DCRL (Full)}  & \textbf{79.2}  & 93.3  & \textbf{14.7}  & 56.7  & \textbf{37.9}   & \textbf{56.3} \\
        - w/o Harmonic Election  & 78.7  & \textbf{93.4}  & 14.3  & 56.2  & 37.8   & 56.0 \\
        - w/o Conservative Reward & 78.3  & 93.3  & 12.2  & \textbf{57.2}  & 36.2   & 55.4 \\
        - w/o Dynamic Sampling   & 76.4  & 90.3  & 14.3  & 52.0  & 34.4  & 53.4 \\
        \bottomrule
    \end{tabular}
    \caption{\label{tab:ablation_study} Ablation Studies to Analyze the Contribution of DCRL Core Modules with the Qwen3-8B-Base Model.}
\end{table*}

To understand the contribution of each component in our DCRL framework, we conduct a series of ablation studies on the Qwen3-8B-Base model, and the results are summarized in Table \ref{tab:ablation_study}. More detailed results are shown in Appendix \ref{sec:implementation details}.   

\paragraph{Impact of Harmonic Election}
Replacing harmonic mean consensus with simple majority voting from the anchor model alone leads to performance drops. This confirms that harmonic election effectively mitigates spurious majority bias by fusing both dominant and diverse exploratory signals to produce more reliable pseudo-labels.

\paragraph{Impact of Conservative Reward}
Simplifying our reward design to a binary scheme (1 for correct, 0 otherwise) results in performance degradation, especially in difficult tasks. This demonstrates that our conservative reward, which reserves a modest reward for anchor majority answers, stabilizes training by preventing extreme fluctuations in advantage estimation and avoiding harsh penalties to high-confidence trajectories.

\paragraph{Impact of Dynamic Sampling}
Using both anchor and explorer samples for training at all times leads to the worst overall performance. This underscores the importance of dynamic sampling in balancing exploration and exploitation: it excludes noisy signals early on to avoid reward hacking and incorporates high-quality exploration later, ensuring stable training while preserving the ability to escape suboptimal modes.

\section{Related Works}
\paragraph{Unsupervised RL for LLMs:} 
LLMs can achieve self-improvement without labeled data via two typical unsupervised RL paradigms: determinism-based methods \cite{prabhudesai2025maximizing, zhang2025right} encourage low-entropy and high-confidence predictions for performance sharpening, while aggregation-based methods \cite{zhang2025corewarding, yu2025restrain, wu2025spine} assign rewards by cross-sample agreement, which takes cross-sample consistency as the proxy of prediction correctness.

\paragraph{Test-time Adaptation for LLMs:} Recent works \cite{akyurek2025surprising} have demonstrated that LLMs can leverage reinforcement learning to conduct test-time adaptation \cite{sun2020testtime}, which effectively enhances model performance on unseen data and even surpasses the performance of standard training protocols. This paradigm \cite{zuo2025ttrl,wu2025spine,liu2025ettrl,zhou2025evolving, zhang2025aqattrl} empowers models to dynamically adapt to novel task distributions without access to extra labeled training data.

\section{Conclusion}

In this paper, we propose DCRL, an unsupervised Reinforcement Learning with Verifiable Rewards (RLVR) framework that transforms intrinsic model robustness into reliable learning signals, enabling LLMs to self-improve on reasoning tasks without annotated data. By (i) adopting an Unlearn-Then-Explore strategy to break dominant suboptimal reasoning patterns and enhance exploration capability, (ii) leveraging a Harmonic Election mechanism to balance reliability and diversity for robust pseudo-label estimation, and (iii) introducing Adaptive Sampling to dynamically regulate the exploration-exploitation trade-off during training, DCRL effectively mitigates the spurious majority bias—a critical limitation of existing label-free RLVR methods. Empirically, extensive evaluations across diverse LLMs and challenging reasoning benchmarks demonstrate that DCRL consistently outperforms current determinism-based approaches and aggregation-based approaches, which paves a scalable path for LLM self-improvement without external supervision.

\section*{Limitations}
Although DCRL successfully mitigates spurious majority issues and boosts reasoning performance via dual consensus and enhanced exploration, it still has key limitations when encountering severe systematic prior bias, where both anchor and explorer signals converge to consistent spurious consensus. This provides little corrective supervision and even reinforces misleading reasoning patterns through policy optimization. Moreover, its performance gains diminish for extremely complex out-of-distribution reasoning tasks that deviate far from the model’s pretraining distribution, as \textit{anchor-explorer} fails to reconstruct novel reasoning paths and dual consensus signals become unreliable.

\normalem 
\nocite{*} 
\bibliography{acl_latex}

\appendix
\section{Implementation Details}
\label{sec:implementation details}
\subsection{Prompt}
We use the same suffix prompt both in the training and evaluation of our experiments to promote clear and step-by-step reasoning:\\
\noindent
\begin{mdframed}[
    backgroundcolor=gray!10,
    linecolor=gray!50,
    linewidth=0.5pt, 
    innerleftmargin=8pt,
    innerrightmargin=8pt,
    innertopmargin=4pt,
    innerbottommargin=4pt
]
\texttt{\textbackslash nPlease reason step by step, and put your final answer within \textbackslash boxed\string{\string}}.
\end{mdframed}
\subsection{Hyperparameters}
Hyperparameter settings of our experiment on Qwen3-8B-Base is shown in Table \ref{tab:hyperparameters}.
\begin{table}[H]
\centering
\renewcommand\arraystretch{1.2}
\caption{Hyperparameter Settings for DCRL Framework.}
\label{tab:hyperparameters}
\begin{tabularx}{\linewidth}{@{}>{\centering\arraybackslash}X >{\centering\arraybackslash}X@{}}
\toprule
\textbf{Hyperparameter} & \textbf{Value} \\
\midrule
Batch size & 128 \\
Mini batch size & 128 \\
Micro batch size & 8 \\
Max prompt length & 4096 \\
Max response length & 3072 \\
Learning rate & $1\times10^{-6}$ \\
LR warmup steps ratio & 0.1 \\
Learning rate warmup & cosine \\
Optimizer & Adam \\
Temperature & 0.6 \\
Top k & -1 \\
Top p & 0.95 \\
Unlearn LR & $3\times10^{-7}$ \\
Number of anchor samples per example & 16 \\
Number of explorer samples per example & 16 \\
Number of samples per example for policy update & 16 \\
Number of samples per example for Testing & 16 \\
Use KL loss & False \\
\bottomrule
\end{tabularx}
\end{table}
\renewcommand\arraystretch{1.0}

\subsection{Baseline Implementation}
For all baselines, we use the official code provided in their public repositories. For TTRL, we set the learning rate to $1 \times 10^{-6}$ and the warm-up ratio to 0.1 for large-scale unsupervised learning. For Co-Rewarding-I, we adopt the DAPO-Math-14k dataset rephrased by Qwen3-32B as provided in the original source code. Besides, no external models are used in all baseline experiments.
All other hyperparameter settings for the baseline are kept identical to the original configuration.

\section{Detailed Results}
% \subsection{Generalization of DCRL Models in Test-time Adaptation }
\subsection{Detailed Results of Ablation Studies}
\label{sec:Ablation Details}
\begin{figure}[h]
  \centering
  \includegraphics[width=0.45\textwidth]{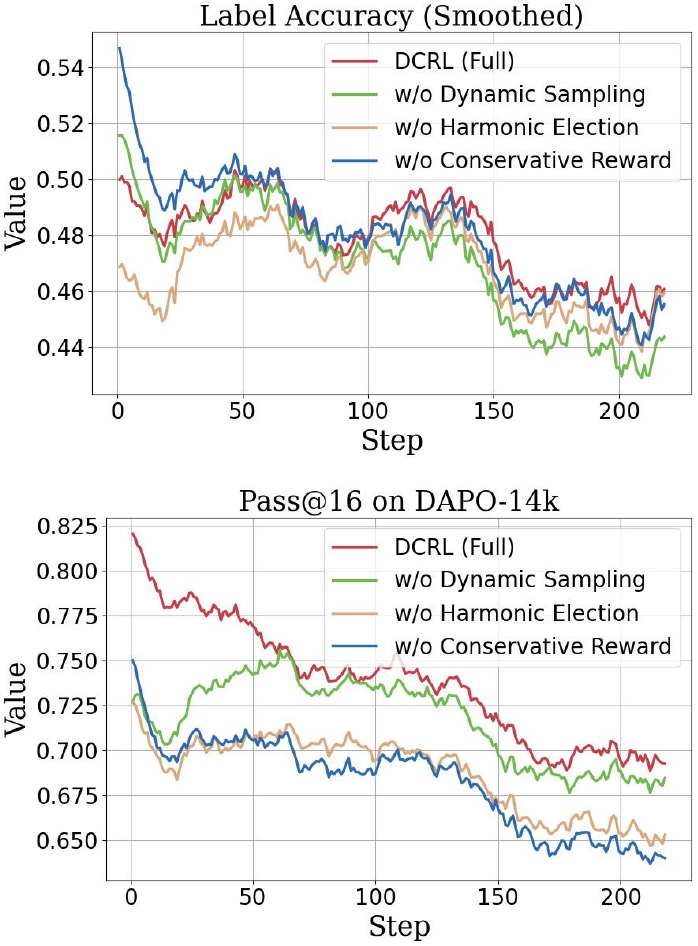}
  \caption{Detailed Results of Ablation Studies with Qwen3-8B-Base, including pass@16 on training dataset and label accuracy}
  \label{fig:detailed_ablation}
\end{figure}

Detailed results of ablation studies are shown in fig \ref{fig:detailed_ablation}.

\section{Extra Experiments}
\subsection{Hyperparameter Sensitivity}
A key empirical finding from our experiments is that differing models necessitate tailored unlearning learning rates ($\eta_{\text{u}}$). A value that is too large risks disrupting the model's ability to generate valid reasoning trajectories, while one that is too small fails to effectively suppress spurious dominant modes. In this section, we present the detailed performance of the Qwen3-8B-Base model under various unlearning learning rate configurations, thereby validating the robustness of our Unlearn-Then-Explore strategy and the rationale behind our specific hyperparameter selection. Results are shown in Table \ref{tab:unlearn_lr_sensitivity}.

\begin{table*}[h]
    \centering
    \small
    \setlength{\tabcolsep}{4.2mm}
    \renewcommand{\arraystretch}{1.05}
    \begin{tabular}{l|ccccc|c}
        \toprule[0.8pt]
        \textbf{Unlearn LR ($\eta_{\text{u}}$)} & \textbf{MATH} & \textbf{GSM8K} & \textbf{AIME24} & \textbf{MMLU.} & \textbf{GPQA.} & \textbf{Average} \\
        \midrule
        $1 \times 10^{-7}$ & 78.6 & 92.7 & 12.6 & 55.5 & 37.2 & 55.3 \\
        $3 \times 10^{-7}$ & \textbf{79.2} & \textbf{93.3} & \textbf{14.7} & \textbf{56.7} & \textbf{37.9} & \textbf{56.3} \\
        $5 \times 10^{-7}$ & 77.6 & 92.7 & 11.5 & 54.3 & 36.4 & 54.5 \\
        $1 \times 10^{-6}$ (Failed) & 67.3 & 88.1 & 12.1 & 50.2 & 35.1 & 50.5 \\
        \bottomrule
    \end{tabular}
    \caption{\label{tab:unlearn_lr_sensitivity} Sensitivity Analysis of Unlearning Learning Rate ($\eta_{\text{u}}$) on Qwen3-8B-Base.}
\end{table*}

\subsection{Comparison of Different Consensus Strategies}
\label{sec:consensus_comparison}

To further validate the effectiveness of our proposed Harmonic Election mechanism, we compare it against several alternative consensus strategies. These strategies differ in how they aggregate the signals from the \textit{anchor} and \textit{explorer} models to select the final pseudo-label $y^*$. The key insight is that a valid reasoning path should be robust, i.e., supported by both the dominant mode (anchor) and the exploratory distribution (explorer).

\begin{figure}[h]
  \centering
  \includegraphics[width=0.45\textwidth]{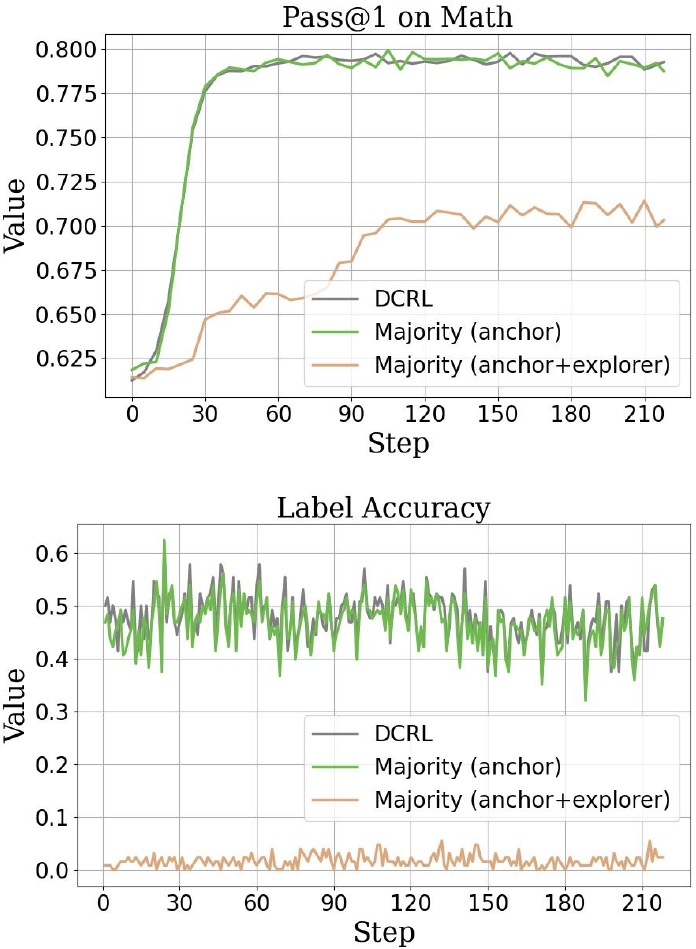}
  \caption{Curves of Different Consensus Strategies for Pseudo-Label Selection on Qwen3-8B-Base.}
  \label{fig:Different_Consensus}
\end{figure}

We evaluate the following strategies:
\begin{itemize}
    \item \textbf{Majority Vote (Anchor Only)}: Simply select the majority answer from the \textit{anchor} model's rollouts.
    \item \textbf{Majority Vote (Anchor + Explorer)}: A simple aggregation strategy that combines all rollouts from both the \textit{anchor} and \textit{explorer} models and selects the majority answer.
    \item \textbf{Harmonic Mean (Ours)}: Our proposed strategy, which selects the answer that maximizes the harmonic mean of its probabilities in the \textit{anchor} and \textit{explorer} distributions.
\end{itemize}
The results are presented in Table \ref{tab:consensus_comparison} and Fig \ref{fig:Different_Consensus}. We observe that simply combining all samples (\textit{Anchor + Explorer}) does not improve performance and can even be detrimental, as it does not effectively filter out spurious signals. In contrast, our harmonic mean strategy achieves the best overall performance, demonstrating its superior ability to balance reliability and diversity in pseudo-label selection.
\begin{table*}[h]
    \centering
    \small
    \setlength{\tabcolsep}{4.0mm}
    \renewcommand{\arraystretch}{1.05}
    \begin{tabular}{l|ccccc|c}
        \toprule[0.8pt]
        \textbf{Consensus Strategy} & \textbf{MATH} & \textbf{GSM8K} & \textbf{AIME24} & \textbf{MMLU.} & \textbf{GPQA.} & \textbf{Avg.} \\
        \midrule
        Majority Vote (Anchor Only) & 78.7  & \textbf{93.4}  & 14.3  & 56.2  & 37.8   & 56.0 \\
        Majority Vote (Anchor + Explorer) & 64.7 & 90.3 & 14.3 & 52.0 & 34.4 & 51.5 \\
        Harmonic Mean (Ours) & \textbf{79.2} & 93.3 & \textbf{14.7} & \textbf{56.7} & \textbf{37.9} & \textbf{56.3} \\
        \bottomrule
    \end{tabular}
    \caption{\label{tab:consensus_comparison} Comparison of Different Consensus Strategies for Pseudo-Label Selection on Qwen3-8B-Base.}
\end{table*}

\section{Why Does Dual Consensus Work?}
\label{app:proof}

We formally prove that the dual consensus pseudo-label selection mechanism achieves higher accuracy than naive majority vote by mitigating spurious majority bias, under mild and realistic assumptions.

\subsection{Problem Setup \& Definitions}
Let $\mathcal{A}$ be the set of candidate answers, and $y_{\text{true}} \in \mathcal{A}$ be the ground-truth answer.
\begin{itemize}[leftmargin=*, itemsep=0pt, topsep=0pt]
    \item $\pi_{\text{anchor}}(a)$: Probability of answer $a$ from the anchor model.
    \item $\pi_{\text{explorer}}(a)$: Probability of answer $a$ from the explorer model (after unlearning).
    \item $p_0(a), p_1(a)$: Empirical probabilities of $a$ from $G$ rollouts of the anchor and explorer models, respectively.
    \item \textbf{Majority Vote}: $\hat{y}_{\text{MV}} = \arg\max_{a} p_0(a)$.
    \item \textbf{Dual Consensus}: $y_{\text{DC}}^* = \arg\max_{a} S(a)$, where $S(a) = \frac{2p_0(a)p_1(a)}{p_0(a)+p_1(a)}$ is the harmonic mean score.
\end{itemize}

\subsection{Key Assumptions}
We introduce three realistic assumptions for LLMs with spurious majority bias.

\noindent\textbf{Assumption 1 (Spurious Majority Bias):}
There exists a spurious dominant answer $y_{\text{sp}} \neq y_{\text{true}}$ such that:
$$
\pi_{\text{anchor}}(y_{\text{sp}}) \gg \pi_{\text{anchor}}(y_{\text{true}})
$$
This is the core failure mode of majority vote.

\noindent\textbf{Assumption 2 (Effective Unlearning):}
The explorer model suppresses the spurious answer but preserves the true answer:
\begin{align*}
\pi_{\text{explorer}}(y_{\text{sp}}) \ll \pi_{\text{anchor}}(y_{\text{sp}}), \\
\quad \frac{\pi_{\text{explorer}}(y_{\text{true}})}{\pi_{\text{anchor}}(y_{\text{true}})} \gg \frac{\pi_{\text{explorer}}(y_{\text{sp}})}{\pi_{\text{anchor}}(y_{\text{sp}})}
\end{align*}
The ratio inequality implies the true answer is more robust to unlearning.

\noindent\textbf{Assumption 3 (Large-Sample Consistency):}
For sufficiently large $G$, by the Law of Large Numbers:
$$
p_0(a) \xrightarrow{G\to\infty} \pi_{\text{anchor}}(a), \quad p_1(a) \xrightarrow{G\to\infty} \pi_{\text{explorer}}(a)
$$

\subsection{Main Result}
\textbf{Theorem}: Under Assumptions 1-3, DCRL selects the true answer ($y_{\text{DC}}^* = y_{\text{true}}$), while majority vote selects the spurious answer ($\hat{y}_{\text{MV}} = y_{\text{sp}}$). Thus, $\text{Acc}(y_{\text{DC}}^*) > \text{Acc}(\hat{y}_{\text{MV}})$.

\subsection{Proof}
\paragraph{Part 1: Majority Vote Converges to Spurious Answer}: By Assumption 1, $\pi_{\text{anchor}}(y_{\text{sp}}) \gg \pi_{\text{anchor}}(y_{\text{true}})$.
By Assumption 3, $p_0(y_{\text{sp}}) > p_0(a),\forall a\in\mathcal{A}$.
Thus $\hat{y}_{\text{MV}} = \arg\max_{a} p_0(a) = y_{\text{sp}}$.

\paragraph{Part 2: Dual Consensus Converges to True Answer}: We show $S(y_{\text{true}}) > S(y_{\text{sp}})$. For large $G$:

$$S(a)\to\tilde{S}(a) = \frac{2\pi_{\text{anchor}}(a)\pi_{\text{explorer}}(a)}{\pi_{\text{anchor}}(a)+\pi_{\text{explorer}}(a)}.$$

Define robustness ratios (Assumption 2, $r_{\text{true}} \gg r_{\text{sp}}$):
\begin{align*}
r_{\text{sp}} &= \frac{\pi_{\text{explorer}}(y_{\text{sp}})}{\pi_{\text{anchor}}(y_{\text{sp}})}, \quad
r_{\text{true}} = \frac{\pi_{\text{explorer}}(y_{\text{true}})}{\pi_{\text{anchor}}(y_{\text{true}})}, \\
&\text{where } r_{\text{true}} \gg r_{\text{sp}} \to 0.
\end{align*}

Substitute ratios into $\tilde{S}(a)$:
\begin{align*}
\tilde{S}(y_{\text{sp}}) &= \frac{2\pi_{\text{anchor}}(y_{\text{sp}})\cdot r_{\text{sp}}\pi_{\text{anchor}}(y_{\text{sp}})}{\pi_{\text{anchor}}(y_{\text{sp}})+r_{\text{sp}}\pi_{\text{anchor}}(y_{\text{sp}})} \\
&= \frac{2r_{\text{sp}} \cdot \pi_{\text{anchor}}(y_{\text{sp}})}{1 + r_{\text{sp}}}.
\end{align*}
\begin{align*}
\tilde{S}(y_{\text{true}}) &= \frac{2\pi_{\text{anchor}}(y_{\text{true}})\cdot r_{\text{true}}\pi_{\text{anchor}}(y_{\text{true}})}{\pi_{\text{anchor}}(y_{\text{true}})+r_{\text{true}}\pi_{\text{anchor}}(y_{\text{true}})} \\
&= \frac{2r_{\text{true}} \cdot \pi_{\text{anchor}}(y_{\text{true}})}{1 + r_{\text{true}}}.
\end{align*}

By Assumption 2, $r_{\text{sp}}\to0 \implies \tilde{S}(y_{\text{sp}})\to0$.
Since $\pi_{\text{anchor}}(y_{\text{true}})>0$ and $r_{\text{true}}>0$, we have $\tilde{S}(y_{\text{true}}) > 0$.
Thus $\tilde{S}(y_{\text{true}}) > \tilde{S}(y_{\text{sp}})$, so $y_{\text{DC}}^* = \arg\max_{a} S(a) = y_{\text{true}}$.
\\
\\
Dual Consensus enforces a robustness constraint—valid answers must be supported by both anchor and explorer, eliminating spurious answers fragile to unlearning and outperforming Majority Vote.

\section{The Use of Large Language Models}
We used large language models (LLMs) only to polish writing  and improve textual clarity. No LLM was applied to research idea generation, experimental design, data analysis, or result derivation. All scientific contributions—conceptualization, methodology, experiments, and conclusions—were independently developed by the authors in full.
\end{document}